\documentclass[10pt,twocolumn,letterpaper,fleqn]{article}

\usepackage{cvpr}
\usepackage{times}
\usepackage{epsfig}
\usepackage{graphicx}
\usepackage{amsmath}
\usepackage{amssymb}
\usepackage{booktabs}
\usepackage{mathrsfs}
\usepackage{algorithm}
\usepackage{algorithmic}
\usepackage{multirow}

\newcommand{\INPUT}{\item[\myinput]}
\newcommand{\myinput}{\textbf{Initialization:}}

\newcommand{\MYWHILE}{\item[\mywhile]}
\newcommand{\mywhile}{\textbf{repeat}}

\newcommand{\MYENDWHILE}{\item[\myendwhile]}
\newcommand{\myendwhile}{\textbf{until}}

\usepackage[pagebackref=true,colorlinks,bookmarks=false]{hyperref}

\cvprfinalcopy 

\def\cvprPaperID{870} 

\ifcvprfinal\fi
\begin{document}

\thinmuskip=0mu
\medmuskip=0mu
\thickmuskip=0mu

\title{Learning Contour-Fragment-based Shape Model with And-Or Tree Representation}

\author{Liang Lin$^{1,2}$, ~~Xiaolong Wang$^{1}$,~~ Wei Yang$^{1}$, ~~ Jianhuang Lai$^{1}$\thanks{This work was supported by the National Natural Science Foundation of China (Grant No.90920009 and No.60970156), the Hi-Tech Research and Development Program of China (National 863 Program, Grant
No.2012AA011504), and the Guangdong Natural Science Foundation (Grant
No.S2011010001378). This work was also supported in part by the Guangdong Science and Technology Program (Grant No.2011B040300029). Corresponding author is Jianhuang Lai.}\\
$^1$Sun Yat-Sen University, Guangzhou, China\\
$^2$Lotus Hill Research Institute, China\\
{\tt\small linliang@ieee.org,   ~~~~stsljh@mail.sysu.edu.cn}
}

\maketitle

\begin{abstract}

This paper proposes a simple yet effective method to learn the hierarchical object shape model consisting of local contour fragments, which represents a category of shapes in the form of an And-Or tree. This model extends the traditional hierarchical tree structures by introducing the ``switch'' variables (i.e. the or-nodes) that explicitly specify production rules to capture shape variations. We thus define the model with three layers: the leaf-nodes for detecting local contour fragments, the or-nodes specifying selection of leaf-nodes, and the root-node encoding the holistic distortion. In the training stage, for optimization of the And-Or tree learning, we extend the concave-convex procedure (CCCP) by embedding the structural clustering during the iterative learning steps. The inference of shape detection is consistent with the model optimization, which integrates the local testings via the leaf-nodes and or-nodes with the global verification via the root-node.  The advantages of our approach are validated on the challenging shape databases (i.e., ETHZ and INRIA Horse) and summarized as follows. (1) The proposed method is able to accurately localize shape contours against unreliable edge detection and edge tracing. (2) The And-Or tree model enables us to well capture the intraclass variance.



\end{abstract}


\vspace{-2mm}
\section{Introduction}
\vspace{-1mm}

\begin{figure}[!htb]
\centering
\epsfig{figure=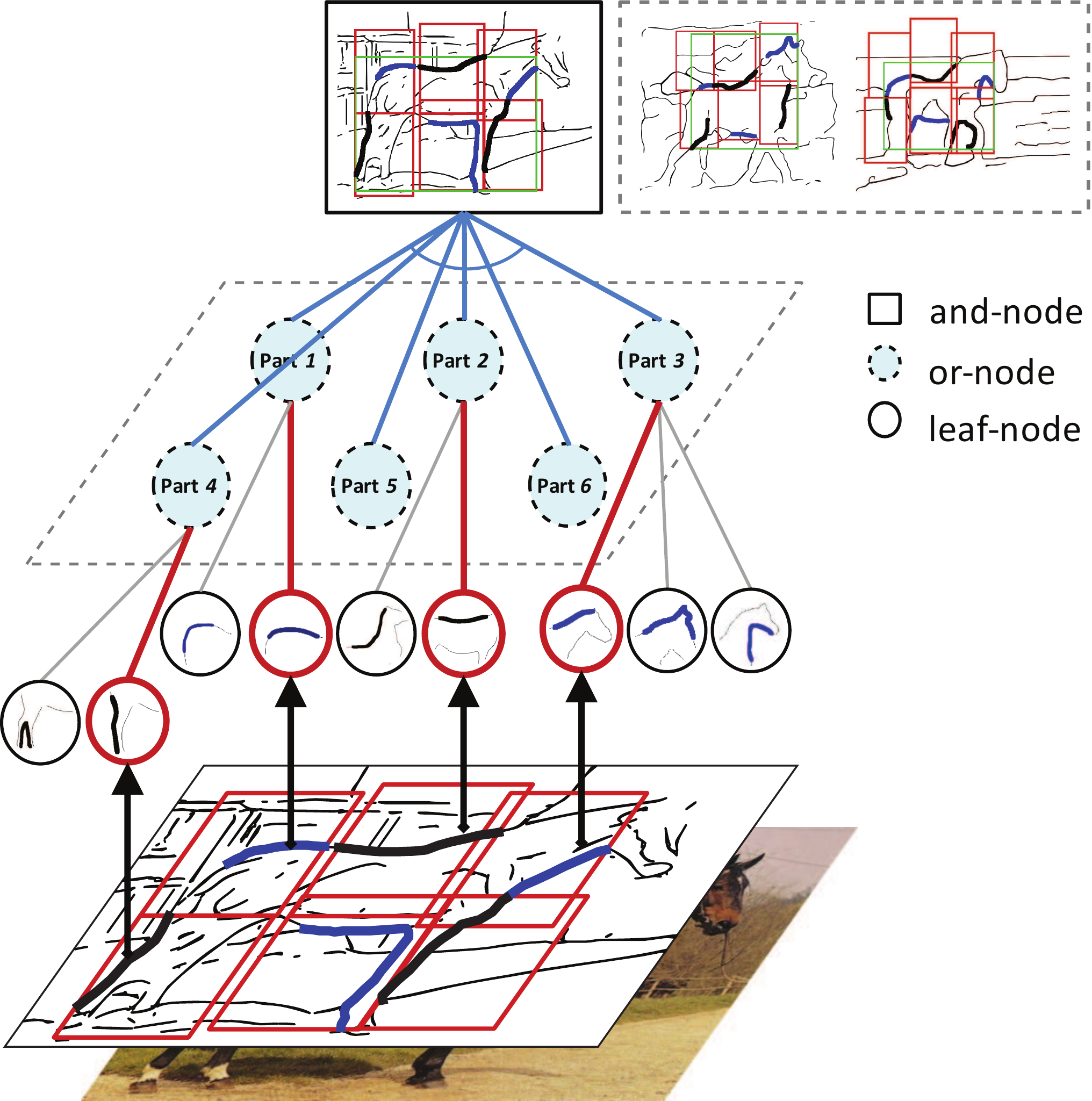,width=3.4in}
\vspace{-5mm}\caption{ An example of shape model in the form of an And-Or tree. It comprises three layers: the leaf-nodes for detecting local contour fragments, the or-nodes specifying selection of leaf-nodes, and the root-node encoding the holistic variance. The bold red vertical lines represent the selection of leaf-nodes in the inference. }\label{fig:AOTree_example}
\end{figure}

Detecting and localizing object shapes from images are areas of active research. This paper studies a novel shape detection method by learning the contour-fragment-based shape model. We represent a category of shapes in the form of a hierarchical And-Or tree, which can be automatically learned in a semi-supervised manner. Fig. ~\ref{fig:AOTree_example} shows an example of learned shape model for horses, consisting of three layers. The bottom layer of the tree includes a batch of leaf-nodes, i.e., the local classifiers used for localizing contour fragments. The middle layer contains a set of or-nodes, each of which explicitly represents a part of the shape and specifies a few leaf-nodes for selection; intuitively, one or-node can be viewed as a ``switch'' variable to activate only one leaf-node at a time during inference. The top root-node (i.e. the and-node) is a global classifier encoding the holistic variance and distortion.

{\em Literature review.} We review the related work in two aspects: shape (or contour) matching and shape model learning.

(i) Many methods pose the shape detection as a task of matching contours in images, and they basically utilize hand-drawn shapes as reference templates~\cite{ShapeContext,ShapeTree,ShiShapeECCV2008,LinGraphMatch,ShapeGroup,PartialMatchingECCV2010,LateckiCVPR2011}. To overcome the difficulties of occlusions (i.e. missing of true contours of objects) and incomplete (broken) contours, a number of robust shape descriptors are extensively discussed such as Shape Context and its extensions~\cite{ShapeContext,LateckiCVPR2011,ShiShapeECCV2008}, Contour Flexiblity~\cite{ConFlexibility}, Local-angle~\cite{PartialMatchingECCV2010,ShapeGroup}, as well as effective matching schemes, e.g., particle filtering~\cite{ShapeGroup}, dynamic programming~\cite{ShapeTree}, stochastic sampling~\cite{LinGraphMatch}. For example, Zhu et al.~\cite{ShiShapeECCV2008} proposed to achieve many-to-many matching of contours by using voting scheme. Riemenschneider et al.~\cite{PartialMatchingECCV2010} addressed partial shape matching by identifying matches from fragments of arbitrary length to the reference contour.




(ii) An alternative to shape detection is achieved by learning shape models for a given category of shape instances. These methods represent shapes as a loose collection of local contour fragments or an ensemble of pairwise constraints~\cite{ShottonPAMI08,BaiShapeBand,LinPR,ShiShapeCVPR2010}. They usually construct a codebook of fragments (e.g., PAS~\cite{PAS} and salient contours~\cite{ShottonPAMI08,LinICCV07}) and train the model by using the boosting algorithm~\cite{ShottonPAMI08}, SVM~\cite{ShiShapeCVPR2010,PAS}, generative learning~\cite{LinECCV2010} or Hough-style voting~\cite{MalikCVPR2009}. However, some of them are limited to learning with clutter-free shape instances~\cite{DrawingShape2006}, and some assume the shape configurations recurring consistently which often suffer from large intraclass variance (e.g. articulation) or highly inaccurate edge detection. Recently, a state-of-the-art for object detection is achieved by~\cite{LatentSVM}, where a tree-structure latent SVMs model is trained using multi-scale HoG feature. It inspires us to define the tree structure shape model; in addition, we extend the structure by introducing the ``switch'' variables (i.e. the or-nodes) that explicitly specify production rules to capture large shape variations.

The key contributions of this work are as follows. First, we propose the shape model in the form of an And-Or tree that enables us to achieve superior performance compared to the state-of-the-art approaches. Second, a novel optimization algorithm is proposed to learn the model structure and parameters simultaneously in a semi-supervised way. There are four key components in our method.

The {\bf leaf-nodes} in the And-Or tree model represent a set of local classifiers of contour fragments. According to the analysis in~\cite{LateckiCVPR2011}, one of the key challenges in shape detection is that true contours of objects often connect to background clutters due to unreliable edge detection and contour tracing. Therefore, addressing this problem, we design a partial matching scheme that can localize the correct part of the contour with the local classifiers.


The {\bf or-nodes} in the middle of the model (see Fig.~\ref{fig:AOTree_example}) are ``switch'' variables specifying the production rules for leaf-node selection. Once a number of contour fragments are detected and localized via the local classifiers, each or-node is used to select one optimal contour fragment as a part of the shape. The benefits of introducing the or-nodes are obvious~\cite{LinICCV07,LinPR}: they provide alternate ways of composition being significant to address the large intraclass variance. Moreover, we allow the or-nodes to slightly perturb their locations during detection, which accounts for deformation and distortion. In our implementation, we fix the or-nodes in a layout of $2 \times 3$ blocks. As Fig.~\ref{fig:AOTree_example} illustrates, in which each block of or-node is denoted by the red box, our model can capture not only the local variant (e.g. part 2) but also the inconsistency caused by edge missing or broken (e.g. part 3).

The {\bf training of our shape model} is posed as an optimization problem of the And-Or tree learning that integrates structure learning and parameter learning. We present a framework based on the CCCP (concave-convex procedure)~\cite{LeoCCCP} by embedding a clustering step during the iteration, namely the dynamic CCCP (dCCCP).


The {\bf inference of shape detection} is consistent with the optimization of training, including two steps. We first perform the bottom-up testings using the leaf-nodes and or-nodes. A number of candidate contour fragments are thus obtained and some of them are activated via the or-nodes. All the selected contour fragments are then combined together via the root-node for global verification.

The rest of this paper is organized as follows. We first present the shape model with the And-Or tree representation in Section 2, and follow with a description of shape model learning in Section 3. The experimental results and comparisons are exhibited in Section 4. A conclusion is presented in Section 5.

\section{Shape Model with And-Or Tree}
We introduce the shape model with three aspects: (i) a descriptor of shape contour, (ii) the And-Or tree representation, and (iii) the inference with the learnt model.

\vspace{-1mm}
\subsection{Contour descriptor}
\vspace{-1mm}

\begin{figure}[!htb]
\centering
\epsfig{figure=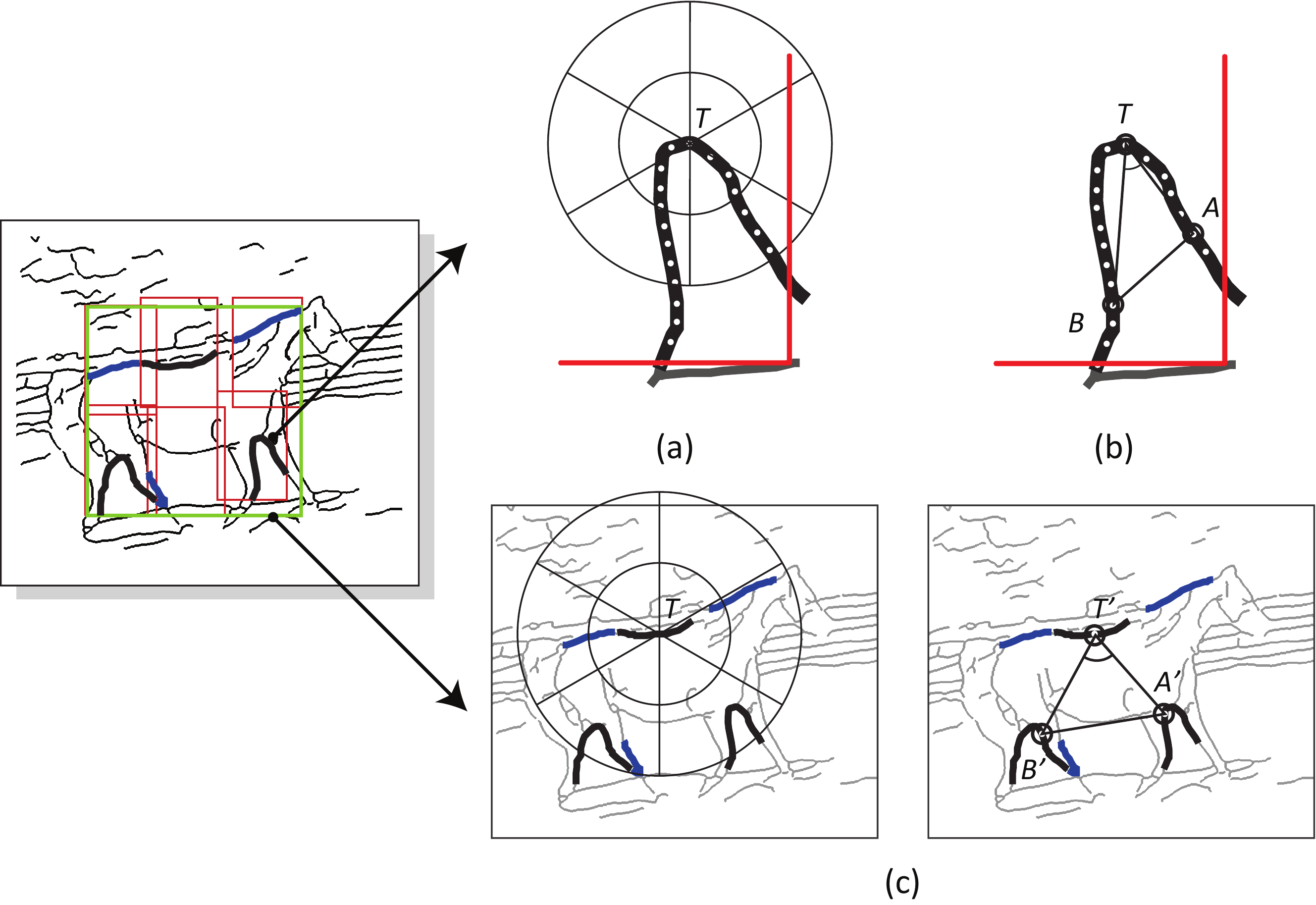,width=3.3in}
\vspace{-5mm}\caption{ Illustration of the contour descriptor. The shape context feature ( in (a) ) and the triangle-based feature ( in (b) ) are computed for a local contour fragment. (c) These two types of features are also suitable for a group of contour fragments representing the shape instance.}\label{fig:feature}
\end{figure}

We start by designing an effective contour descriptor that combines the triangle-based feature proposed in~\cite{ShapeGroup} and Shape Context~\cite{ShapeContext}. As Fig.~\ref{fig:feature} illustrates, this descriptor is suitable for a local contour fragment as well as a group of contour fragments representing an object shape.

To describe a local contour fragment, we first extract a sequence of sample points $\Omega$ from the contour fragment. For each point in $\Omega$, we compute its triangle-based descriptor as well as the Shape Context descriptor. By combining these two types of descriptors for each point in $\Omega$, we obtain a discriminative and deformation-tolerant descriptor for this contour fragment. As for describing an object shape represented by several contour fragments, the points in $\Omega$ are selected from the whole object shape.

Given a point $T \in\Omega$, the triangle-based descriptor is a histogram of all triangles constructed by $T$ and any other two different points $A, B \in \Omega$ (Fig.~\ref{fig:feature}(b)). More precisely, it is a 3D histogram of the angles $BTA$, and two distances $TA$ and $TB$. Note that triangle $BTA$ is oriented clockwise and distances $TB$ and $TA$ are normalized by the average distance between points in $\Omega$. As for the Shape Context descriptor, it considers the lengths and orientations of the vectors from $T$ to all other points in $\Omega$.

In our implementation, the number of sample points in $\Omega$ is fixed to 20, and the distances between adjacent points in $\Omega$ are equal. We represent the 3D histogram with 2 bins for distance $TA$, 2 bins for distance $TB$, and 6 bins for angle $BTA$ ranging from $0$ to $\pi$. We then transform the 3D histogram into a $2 \times 2 \times 6 = 24$-bin 1-D feature vector. As for representing the Shape Context descriptor, we use 2 bins for vector distances, and 6 bins for vector orientations ranging from $0$ to $2\pi$. The length of the 1-D feature vector of Shape Context is $2 \times 6 = 12$ dimensions. By combining these two descriptors as a composite descriptor, the feature vector of the whole point sequence $\Omega$ is the ensemble of the composite descriptor of each point in $\Omega$, with the length $(24+12)\times 20=720$ dimensions.

\vspace{-1mm}
\subsection{And-Or Tree Representation}
\vspace{-1mm}

Our model is defined with three types of nodes in three layers: one root-node (i.e. the and-node), and a number of or-nodes and leaf-nodes described by square, dashed circle and solid circle, respectively, in Fig.~\ref{fig:AOTree_example}. The root and-node represents the whole object, and it has 6 children (or-nodes) in a layout of $2 \times 3$ blocks, each representing one part of a shape. The number of leaf-nodes is unfixed but less than $m$ for each or-node. Assume the maximum number of nodes in the model is $1+n = 1 + 6 + 6 \times m$: $0$ indexes the root node, $i=1,...,6$ indexes the or-nodes and $j=7,...,n$ indexes the leaf-nodes. We also define that $j \in ch(i)$ indexes the child nodes of node $i$. Note that we index $m$ leaf-nodes for each or-node even if some of them do not exist, whose parameters are set as $0$. We present the definitions for the three types of nodes as follows.

\textbf{Leaf-node:} Each leaf-node $L_j,j=7,...,n$ is one classifier of local contour fragment corresponding to its parent node. All leaf-nodes belonging to the same or-nodes (the localized block) have the same location in the image. Suppose the location of the block is $p_i = (p_i^x,p_i^y)$ and a contour fragment $c_j$ is selected as input of the classifier. We denote $\phi^{l}(p_i,c_j)$ as the feature vector using the contour descriptor. Note that only the part of $c_j$ inside the block is taken into account, as Fig.~\ref{fig:feature} illustrates. If the contour $c_j$ is entirely out of the block, $\phi^{l}(p_i,c_j) = 0$. Therefore, the response of classifier $L_j$ for $c_j$ in location $p_i$ is defined as:
\begin{eqnarray}
&& \mathcal{R}_{j}(p_i,c_j) = \omega_{j}^{l} \cdot \phi^{l}(p_i,c_j),
\end{eqnarray}
where $\omega_{j}^{l}$ is a parameter vector. We set $\omega_{j}^{l} = 0$ if the leaf-node $L_j$ is empty.

\textbf{Or-node:} Each or-node $U_i,i=1,...,6$ is used to specify an appropriate contour fragment from a set of detection candidates via its children leaf-nodes $L_j$.

In order to encode the shape deformation, the or-nodes are allowed to perturb slightly with respect to the shape instance. For each or-node $U_i$, we introduce an offset $d_i = (d_i^x, d_i^y)$ to describe the expected spatial position relative to the position of root node $p_0 = (p_0^x, p_0^y)$. Suppose the or-node block is located at $p_i = (p_i^x,p_i^y)$, the difference between $p_i$ and the expected position is $(dx,dy)$, in which $dx = p_i^x - (p_0^x + d_i^x)$ and  $dy = p_i^y - (p_0^y + d_i^y)$. Therefore, given the or-node $U_i$, the cost for the deformation of a leaf-node $L_j$ is defined as:
\begin{eqnarray}
&& Cost_{i,j}(p_0,p_i) = \omega_{j}^{s} \cdot \phi^{s}(p_0,p_i),
\end{eqnarray}
where $\phi^{s}(p_0,p_i)=(dx,dy,dx^2,dy^2)$ is the deformation feature, and $\omega_{j}^{s}$is a 4-dimensional parameter vector for $L_j$, we set $\omega_{j}^{s} = 0$ if $L_j$ is empty.

The advantages are very clear for introducing the or-nodes in the tree structure. (1) The intraclass variance and inconsistency caused by edge computation can be captured by different leaf-nodes specified by the or-nodes. (2) The location flexibility of or-nodes can deal with the non-rigid deformation or local displacement of shapes.

\textbf{Root-node:} The root node at the top is a global classifier for a set of contour fragments $C^r$ proposed by the or-nodes. The response of the root-node is defined similarly with the local classifiers for the leaf-nodes, as:
\begin{eqnarray} \label{eq:root_score}
&& \mathcal{G}(C^r) = \omega^{r} \cdot \phi^{r}(C^r),
\end{eqnarray}
where $\phi^{r}(C^r)$ is the feature vector of $C^r$ and $\omega^{r}$ is the corresponding parameter vector.


\vspace{-1mm}
\subsection{Inference with And-Or Tree}
\vspace{-1mm}

Given the learnt And-Or tree model, the inference task is to localize optimal contour fragments within the detection window. The target shape (i.e. the root-node) is located by sliding the detection window at all positions and scales of the edge map $X$. Assuming the location of the root-node is $p_0 = (p_0^x,p_0^y)$, we describe the inference as follows.

{\small \textbullet} Bottom-up local testing: For each leaf-node $L_j$, assume the block of its parent $U_i$ is located at $p_i$. The detection score of $L_j$ is calculated by selecting a contour fragment with the highest classifier response,
\begin{eqnarray}
S_{L_j}(X,p_i) &=& \max_{c_j \in X} \mathcal{R}_{j}(p_i,c_j) \nonumber\\
&=& \max_{c_j \in X} \omega_{j}^{l} \cdot \phi^{l}(p_i,c_j).
\end{eqnarray}

The detection score of the or-node $U_i$ is calculated by specifying a contour from the candidates localized by its children leaf-nodes. The deformation cost for the block of $U_i$ is taken into account as well. For clear definition, we introduce an auxiliary  ``switch'' vector $\textbf{v}_{i} = (v_{j_1},v_{j_2},...,v_{j_m} )$ , where $v_j \in \{0,1\}$ and $ ||\textbf{v}_{i}|| = 1$, to indicate which contour is chosen from $m$  candidates via $U_i$. Therefore, the score of the or-node is defined as,
\begin{small}
\mathindent=12pt
\begin{align}
& S_{U_i}(X,p_0) = \max_{\textbf{v}_{i},p_i} \sum_{j\in ch(i)}(S_{L_j}(X,p_i) \cdot v_j - Cost_{i,j}(p_0,p_i) \cdot v_j) \nonumber\\
& =  \max_{\textbf{v}_{i},p_i} \sum_{j \in ch(i)} \max_{c_j \in X}(\omega_{j}^{l} \cdot \phi^{l}(p_i,c_j) \cdot v_j -\omega_{j}^{s} \cdot \phi^{s}(p_0,p_i) \cdot v_j )\nonumber\\
& =  \max_{\textbf{v}_{i},p_i,\textbf{c}_{i}} \sum_{j \in ch(i)} (\omega_{j}^{l} \cdot \phi^{l}(p_i,c_j) \cdot v_j -\omega_{j}^{s} \cdot \phi^{s}(p_0,p_i) \cdot v_j ),
\end{align}
\mathindent=\leftmargini
\end{small}
where $\textbf{c}_{i}$ is a vector representing the input contours for each children leaf-node, $\textbf{c}_{i} = ( c_{j_1},c_{j_2},...,c_{j_m})$.

\begin{figure}[!htb]
\centering
\epsfig{figure=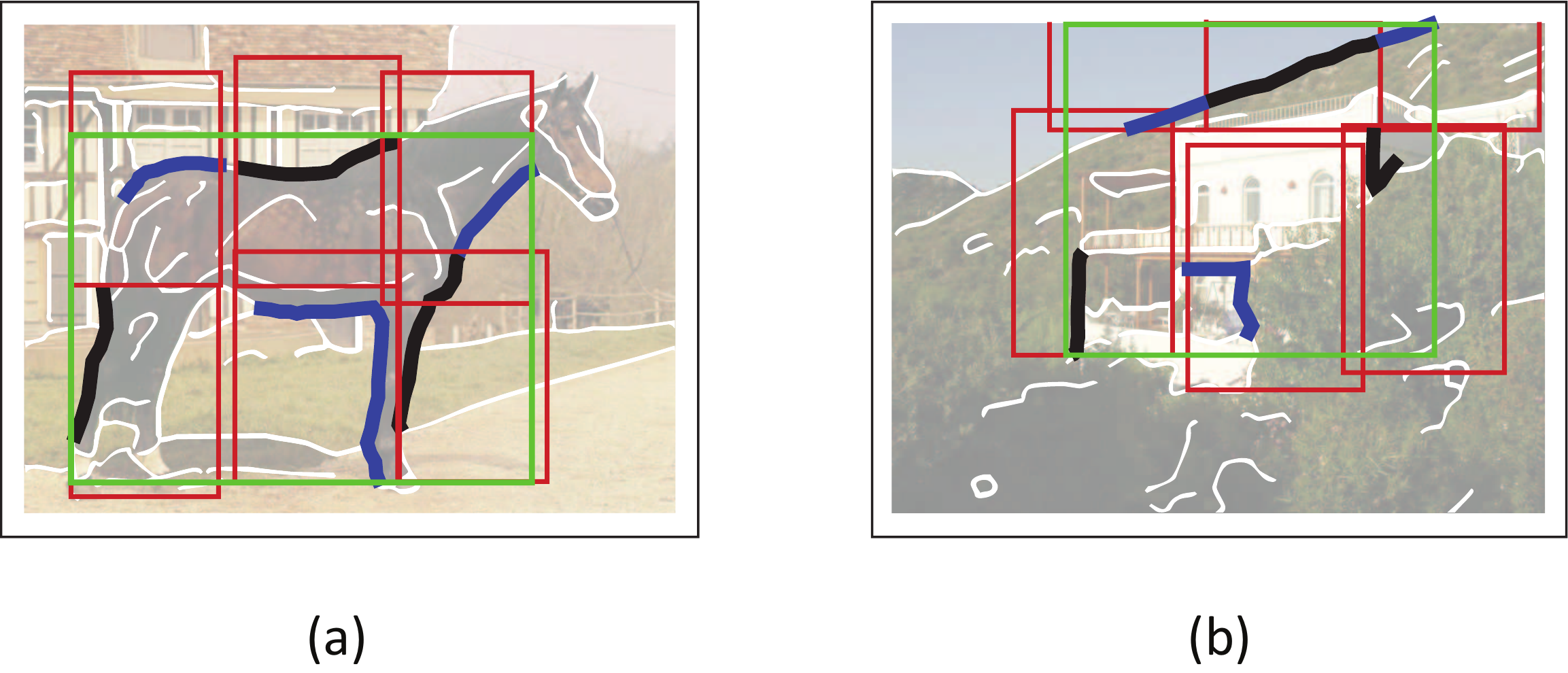,width=3.3in}
\vspace{-5mm}\caption{ Illustration of shape detection. The red boxes denote bottom-up testings with the leaf-nodes and or-nodes, and the green box global verification via the root-node. The two detections ( in (a) and (b) ) have the similar scores of bottom-up testings (i.e., $0.561 : 0.364$) but different scores at the root-node (i.e., $0.093$ : $-0.458$).}\label{fig:inference}
\end{figure}

{\small \textbullet} Verification via the root-node: We obtain a set of contours based on the local proposals, $C^r = \{c_i \}$ where $c_i$ is a contour activated by $U_i$. Then the verification is achieved by calculating the response of root-node defined in Equation(\ref{eq:root_score}). As a result, the overall inference score within the detection window is defined as,
\begin{small}
\mathindent=12pt
\begin{align}\label{eq:score_infer}
& S(X,p_0) = \sum_{i=1}^6 S_{U_i}(X,p_0) + \mathcal{G}(C^r) \nonumber\\
& = \sum_{i=1}^6 \max_{\textbf{v}_i,p_i,\textbf{c}_i} \sum_{j \in ch(i)} (\omega_{j}^l \cdot \phi^l(p_i,c_j) \cdot v_j
- \omega_{j}^{s} \cdot \phi^{s}(p_0,p_i) \cdot v_j )  \nonumber\\
&\quad + \omega^r \cdot \phi^{r}(C^r) \nonumber\\
& =  \max_{V,P,C} \sum_{i=1}^6 \sum_{j \in ch(i)}(\omega_{j}^{l} \cdot \phi^{l}(p_i,c_j) \cdot v_j
- \omega_{j}^{s} \cdot \phi^{s}(p_0,p_i) \cdot v_j )  \nonumber\\
&\quad + \omega^r \cdot \phi^{r}(C^r),
\end{align}
\mathindent=\leftmargini
\end{small}
where $V$ is a joint vector for each $\textbf{v}_i$: $V =(\textbf{v}_1,...,\textbf{v}_6) = (v_7,...,v_n)$, $C$ a joint vector for each $\textbf{c}_i$: $C=(\textbf{c}_1,...,\textbf{c}_6) = (c_7,...,c_n)$ and $P$ a vector of the positions of or-nodes: $P=(p_1,...,p_6)$. We define $H=(V,P,C)$ , and Equation(\ref{eq:score_infer}) can be simplified as,
\begin{eqnarray}\label{eq:sim_score_infer}
&& S(X,p_0) =\max_H \omega \cdot \phi(X,H,p_0),
\end{eqnarray}
where $\omega$ is the vector of model parameters and $\phi(X,H,p_0)$ is the feature vector,
\begin{small}
\mathindent=12pt
\begin{align} \label{abteq65}
& \omega = (\omega_{7}^{l},...,\omega_{n}^{l},-\omega_{7}^{s},...,-\omega_{n}^{s},\omega^{r}). \\
& \phi(X,H,p_0) = (\phi^{l}(p_1,c_7) \cdot v_7,\cdots, \phi^{l}(p_6,c_n) \cdot v_n,  \nonumber\\
& \qquad \qquad \qquad  \phi^{s}(p_0,p_1)\cdot v_7,\cdots,\phi^{s}(p_0,p_6) \cdot v_n, \phi^r(C^r)).
\end{align}
\mathindent=\leftmargini
\end{small}

We present an example to illustrate the inference with the shape model in Fig.~\ref{fig:inference}. The leaf-nodes are used to localize candidate contour fragments and or-nodes to specify the optimal ones; the red boxes denote the results of bottom-up testings. Then we perform the global verification via the root node denoted by the green box, whose significance can be clearly demonstrated in the false positive shown in Fig.~\ref{fig:inference} (b): the aggregation of local similarities needs to be verified.

\vspace{-1mm}
\section{Discriminative Learning for And-Or Tree}
\vspace{-1mm}

The learning of And-Or tree model is an optimization problem that integrates structure learning and parameter learning. The proposed learning framework enables us to learn the structure and the parameters of the model in an alternative way, which is an extension of the original CCCP proposed in~\cite{SVMICML2009}. The significance of this algorithm is as follows. First, we can adjust the layout of parts (decided by the or-nodes) accounting for shape variants within the data. Second, the leaf-nodes can be automatically merged and created to fit the inferred latent variables. More specifically, two leaf-nodes having similar discriminative ability (i.e. to localize similar contours) are encouraged to be merged into one node; one new leaf-node is encouraged to be created for detecting the contours that cannot be handled by current model.

\begin{figure}[!htb]
\centering
\epsfig{figure=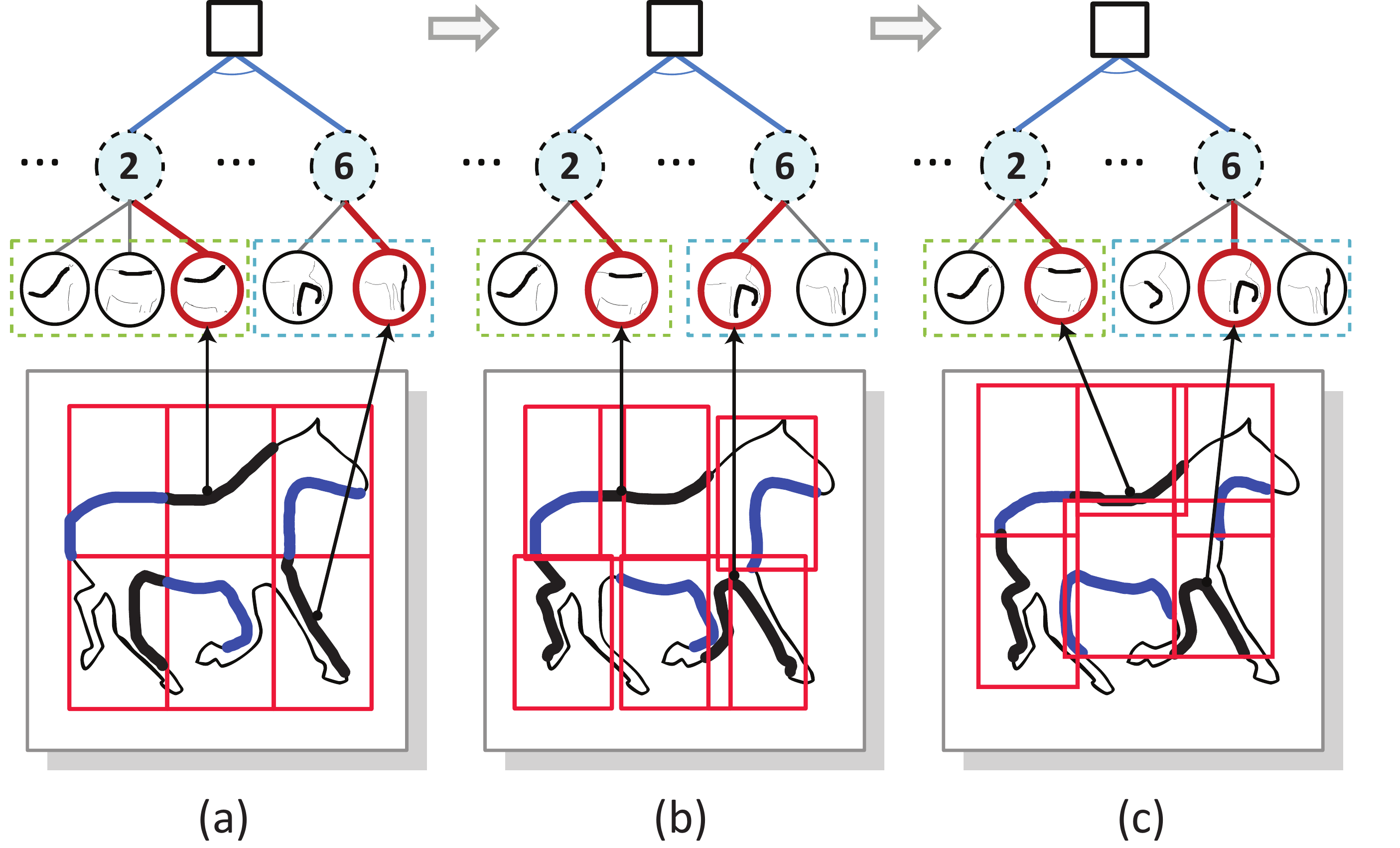,width=3.3in}
\vspace{-5mm}\caption{ Illustration of structure clustering during the learning iterations. We visualize parts of the model in three intermediate steps. Note that each part implies an or-node in the model. (a) The initial structure, i.e. the original regular layout. Two new structures are generated along with the changing of latent variables. (b) Two leaf-nodes belonging to part 2 are merged together. (c) A new leaf-node is created and assigned to part 6. }\label{fig:learning}
\end{figure}


\vspace{-1mm}
\subsection{Optimization Formulation}
\vspace{-1mm}

Given a set of positive and negative training samples $(X_1,y_1)$,...,$(X_N,y_N)$, where $X$ is the edge map within the detection window, $y=\pm 1$ is the label for $X$. We assume the first $K$ samples indexed from $1$ to $K$ are positive samples. Letting $y=1$ denote the positive samples and $y=-1$ the negative samples, we define the feature vector for each sample $(X,y)$ as,
\begin{equation}
 \phi(X,y,H) =
  \left\{
   \begin{array}{lr}
   \phi(X,H) & \mbox{if } y = +1 \\
   0 & \mbox{if } y = -1\\
   \end{array}
  \right.,
\end{equation}
where $H$ is the latent variables for $X$, $\phi(X,H)$ is equivalent to $\phi(X,H,p_0)$, since the position of root-node $p_0$ is fixed. Thus Equation(\ref{eq:sim_score_infer}) can be rewritten as a discriminative function,
\begin{eqnarray} \label{eq:discriminative_fun}
&& S_{\omega}(X) = argmax_{y,H} (\omega \cdot \phi(X,y,H)).
\end{eqnarray}

We can learn the discriminative function(i.e. Equation(\ref{eq:discriminative_fun})) by optimizing the target using structural SVM with latent variables, as,
\begin{small}
\mathindent=0pt
\begin{align} \label{eq:learn_opt}
& \min_{\omega} \frac{1}{2} \| \omega \|^2  +  D\sum_{k=1}^N[\max_{y,H}(\omega \cdot \phi(X_k,y,H) + \mathcal{L}(y_k,y,H)) \nonumber\\
&\qquad - \max_H (\omega \cdot \phi(X_k,y_k,H))],
\end{align}
\mathindent=\leftmargini
\end{small}
where $D$ is a fixed penalty parameter (set as $0.005$ empirically), $\mathcal{L}(y_k,y,H)$ is the loss function. $\mathcal{L}(y_k,y,H) = 0$ if $y_k = y$, else $\mathcal{L}(y_k,y,H) = 1$ in our detection problem.

The optimization target defined in Equation(\ref{eq:learn_opt}) is non-convex. The CCCP framework was recently proposed in~\cite{SVMICML2009,LeoCCCP} to convert it into a convex and concave form and obtain a local optimum solution. Following this framework, we rewrite the target as
\begin{small}
\mathindent=0pt
\begin{align}\label{eq:cccp_f}
&\quad \min_{\omega}[ \frac{1}{2} \|\omega \|^2 + D \sum_{k=1}^N \max_{y,H}(\omega \cdot \phi(X_k,y,H) + \mathcal{L}(y_k,y,H))]  \\ \label{eq:cccp_g}
&\qquad\quad -  [D \sum_{k=1}^N \max_{H} (\omega \cdot \phi(X_k,y_k,H))] \\ \label{eq:opt_target}
& =  \min_{\omega} [f(\omega) - g(\omega)],
\end{align}
\mathindent=\leftmargini
\end{small}
where $f(\omega)$ represents the first two terms in (\ref{eq:cccp_f}), and $g(\omega)$ represents the last term in (\ref{eq:cccp_g}). However, the original CCCP relies on the assumption that the tree structure is fixed during the learning iterations, which is not suitable for our goal, as we need to simultaneously learn the And-Or structures. An extension of CCCP, namely dynamic CCCP (dCCCP) is thus proposed to embed structural clustering into the model parameter learning.

\vspace{-1mm}
\subsection{Optimization with dynamic CCCP}
\vspace{-1mm}

In our learning algorithm, we allow the structure of our model to be dynamically adjusted during each iteration of learning, as Fig.~\ref{fig:learning} illustrates. The proposed dCCCP framework iterates with the following three steps.

{\bf Step 1.} For optimization, we first need to construct a hyperplane that upper bounds the concave part $-g(\omega)$ of the target function. Given the parameter vector $\omega_t$ learned from the previous iteration, we find the hyperplane $q_t$ such that
\begin{equation}\label{eq:CCCP_cons}
 -g(\omega) \leq -g(\omega_t) + (\omega-\omega_t) \cdot q_t, \forall \omega.
\end{equation}
It is performed by searching the best latent variable for each training data $H_k^* = argmax_{H} (\omega_{t} \cdot \phi(X_k,y_k,H))$. Note that $\phi(X_k,y_k,H) = 0$ when $ y_k = -1$, thus we only need to estimate the latent variables for positive training data. Then the hyperplane is constructed as $q_t = - D\sum_{k=1}^N \phi(X_k,y_k,H_k^*)$.

{\bf Step 2.} Given $H_k^* = (V_k^*,P_k^*,C_k^*)$ of all positive samples, the contour fragments can be localized from each positive sample $X_k$. For each or-node $U_i$, we obtain a set of activated contour fragments $\{c_i^{1},c_i^{2},...,c_i^{K}\}$ from all positive samples $\{ X_1, \ldots, X_K\}$. Among this set, we first group the fragments detected via the same leaf-node into the same cluster as a temporary partition, and then apply ISODATA algorithm to perform re-clustering on these contour fragments. Each contour fragment $c_i^k$ is described by the feature $\phi^l(p_i, c_i^k)$ and the Euclidean distance is adopted during the clustering.
The number of clusters are limited to $m$ with regard to the parameter $\omega$. After clustering, each cluster represents a ``potential'' leaf-node whose parameters will be trained in the step 3. We need to decide the new structure in this step and thus assign these potential leaf-nodes to parent or-nodes.

The latent variables $H_k^{d}$ for each positive sample is also changed from $H_k^*$ along with the structure adjustment. Suppose the new hyperplane is $q_t^{d} = - D\sum_{k=1}^N \phi(X_k,y_k,H_k^{d})$. To maintain the property in Equation(\ref{eq:CCCP_cons}), we constrain the newly generated $q_t^{d}$ by $\| q_t - q_t^{d}\| < \varepsilon$ during the clustering procedure, where $\varepsilon$ is set manually. Intuitively, we check the constraints in each step of splitting or merging clusters, which is used to restrict the structure adjustment in an appropriate level.


{\bf Step 3.} Given the latent variables and newly generated structure, the parameters of the model are learned by solving the optimization problem: $\omega_{t+1} = argmin_\omega[f(\omega)+\omega \cdot q_t^{d}]$. By substituting $f(\omega)$ with the first two terms in Equation(\ref{eq:cccp_f}), it can be written as,
\begin{small}
\mathindent=12pt
\begin{align}
& \min_\omega \frac{1}{2} \|\omega\|^2 +  D\sum_{k=1}^N[\max_{y,H}(\omega \cdot \phi(X_k,y,H) + \mathcal{L}(y_k,y,H))  \nonumber\\
&\qquad -  \omega \cdot \phi(X_k,y_k,H_k^d)].
\end{align}
\mathindent=\leftmargini
\end{small}

This is a standard structural SVM problem, let $\Delta\phi(X_k,y,H) = \phi(X_k,y_k,H_k^{d}) - \phi(X_k,y,H)$, the solution can be expressed as,
\begin{eqnarray}
&& \omega^* = D \sum_{k,y,H} \alpha_{k,y,H}^* \Delta\phi(X_k,y,H),
\end{eqnarray}
where $\alpha^*$ can be obtained by maximizing the dual function:
\begin{small}
\mathindent=0pt
\begin{align}
& \max_{\alpha} \sum_{k,y,H} \alpha_{k,y,H} \mathcal{L}(y_k,y,H) \nonumber\\
& - \frac{ D }{2}\sum_{k,k^{\prime}}\ \ \sum_{y,H,y^{\prime},H^{\prime}} \alpha_{k,y,H} \alpha_{k^{\prime},y^{\prime},H^{\prime}} \Delta\phi(X_k,y,H)
\Delta\phi(X_{k^{\prime}},y^{\prime},H^{\prime}),
\end{align}
\mathindent=\leftmargini
\end{small}
which is a dual problem in standard SVM. We solve this problem by applying the cutting plane method~\cite{CuttingPlane} and Sequential Minimal Optimization~\cite{SMO}. Once the parameters $\omega_{t+1}$ is obtained, we repeat the 3-step iteration until the function defined in Equation(\ref{eq:opt_target}) converges.

\vspace{-1mm}
\subsection{Initialization}
\vspace{-1mm}

At the beginning of learning, the block of each or-node is set by regular decomposition, i.e., $(dx,dy) = (0,0)$. Since no leaf-node stands at the beginning, given the positive samples, we select the contours with largest length for each or-node. The structure of our shape model is initialized by clustering without any constrains, and the initial latent variables are obtained accordingly. The parameters of the initialized model can be calculated by solving the standard structural SVM problem.

Algorithm~\ref{alg:Framwork} summarizes the overall algorithm of learning shape model with the And-Or tree.

\begin{small}
\begin{algorithm}[htb]
\caption{Learning Shape Model of the And-Or tree.}
\label{alg:Framwork}
\begin{algorithmic}\footnotesize
\REQUIRE ~~\\
    positive and negative training samples, \\
    $\{X_k,y_k\}^{+} \bigcup \{X_{k^{\prime}},y_{k^{\prime}}\}^{-}, k = 1..K, k^{\prime} = K+1..N$.
\ENSURE ~~\\                           
    The structure and parameters of the shape model.

\INPUT ~~\\
\begin{itemize}
\setlength{\itemsep}{1pt}
 \setlength{\parskip}{0pt}
 \setlength{\parsep}{10pt}

  \item[1] Initialize the structure of model and the latent variables.
  \item[2] Initialize the parameters of model.
\end{itemize}

\MYWHILE
    \STATE
    \begin{itemize}
\setlength{\itemsep}{1pt}
 \setlength{\parskip}{0pt}
 \setlength{\parsep}{10pt}
      \item[1] Estimate the latent variables $H$ by applying inference on each positive sample $(X_k,y_k)$ with the current model.
      \item[2]
        \begin{itemize}
\setlength{\itemsep}{1pt}
 \setlength{\parskip}{0pt}
 \setlength{\parsep}{10pt}
           \item[(a)] Localize the contour fragments for each sample $(X_k,y_k)$ using the current latent variables $H_k^{*}$.
           \item[(b)] For each or-node $U_i$, apply the clustering algorithm with constrains on the contours $\{c_i^{1},c_i^{2},...,c_i^{K}\}$ localized in the same block.
           \item[(c)] Explore a new structure by re-assigning leaf-nodes with or-nodes and modifying the latent variables for each sample from $H_k^*$ to $H_k^d$.
         \end{itemize}

      \item[3] Estimate the model parameters $\omega$ with the fixed model structure and latent variables $H^d$.
    \end{itemize}
\MYENDWHILE {The target function defined in Equation(\ref{eq:opt_target}) converges.}

\end{algorithmic}
\end{algorithm}
\end{small}

\vspace{-1mm}
\section{Experiments}
\vspace{-1mm}

We apply the proposed method on shape detection from images, using the ETHZ database~\cite{PAS} and the INRIA-Horse database~\cite{INRIAHorse} for validation.


\begin{figure}[!htb]
\centering
\epsfig{figure=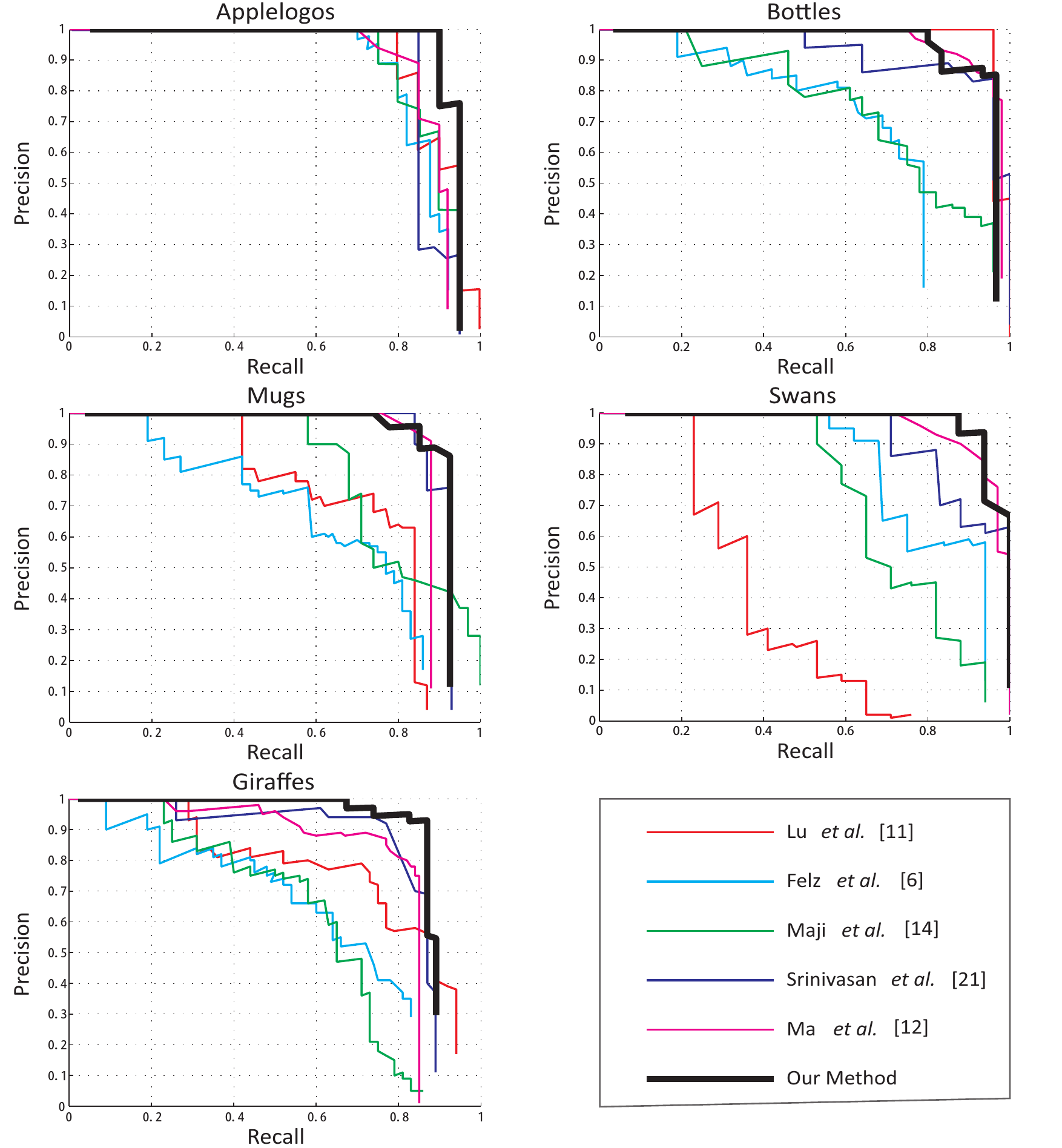,width=3.5in}
\vspace{-5mm}\caption{The precision-recall curves on the ETHZ database. The black (bold) curves represent the results of our method, and the other curves are reported from the previous works. }\label{fig:ETHZ_PR}
\end{figure}

\begin{table*}
  \centering
  \begin{tabular*}{0.71\textwidth}{lcccccc}
    \toprule
     & Applelogos
     & Bottles
     & Giraffes
     & Mugs
     & Swans
     & Average\\
    \hline
    Our method          & \textbf{0.909} & 0.898 & \textbf{0.811} & \textbf{0.893} & \textbf{0.964} & \textbf{0.895} \\
    Ma et al.~\cite{LateckiCVPR2011}           & 0.881 & 0.920 & 0.756 & 0.868 & 0.959 & 0.877 \\
    Srinivasan et al.~\cite{ShiShapeCVPR2010}   & 0.845 & 0.916 & 0.787 & 0.888 & 0.922 & 0.872 \\
    Maji et al.~\cite{MalikCVPR2009}         & 0.869 & 0.724 & 0.742 & 0.806 & 0.716 & 0.771 \\
    Felz et al.~\cite{ShapeTree}         & 0.891 & \textbf{0.950} & 0.608 & 0.721 & 0.391 & 0.712 \\
    Lu et al.~\cite{ShapeGroup}           & 0.844 & 0.641 & 0.617 & 0.643 & 0.798 & 0.709 \\
    \bottomrule
  \end{tabular*}
  \caption{ Quantitative results and comparisons with average precision (AP) on the ETHZ database.}\label{tab:ETHZ_AP}
\end{table*}

{\em Experiment setting.} We fix the number of or-nodes in the shape model as $6$, and the initial layout is a regular partition (e.g. $2 \times 3$ blocks). The maximum number of leaf-nodes for each or-nodes are set as $3$. The shape model training is performed in a semi-supervised manner; the clutter-free contours of positive shapes are labeled and the structures of the models are determined automatically. We extract edge maps for negative examples using the Pb edge detector~\cite{PbDetector} with an edge link algorithm. We adopt PASCAL Challenge criterion as the testing standard: a detection is counted as valid only if the intersection-over-union ratio (IoU) with the groundtruth bounding-box is greater than $50\%$, otherwise detections are counted as false positives. We also submit all the results of shape detection generated by our method in the supplemental material.

The learning algorithm converges after $5 \sim 7$ iterations. During detection, images were searched at 6 different scales, 2 per octave. We carry out the experiments on a PC with Core Duo 3.0 GHZ CPU and 16GB memory. On average, it takes $4 \sim 8$ hours for training a shape model, depending on the numbers of training/testing examples; the time cost for a detection on a image is around $2 \sim 3$ minutes.

\begin{figure}[!htb]
\centering
\epsfig{figure=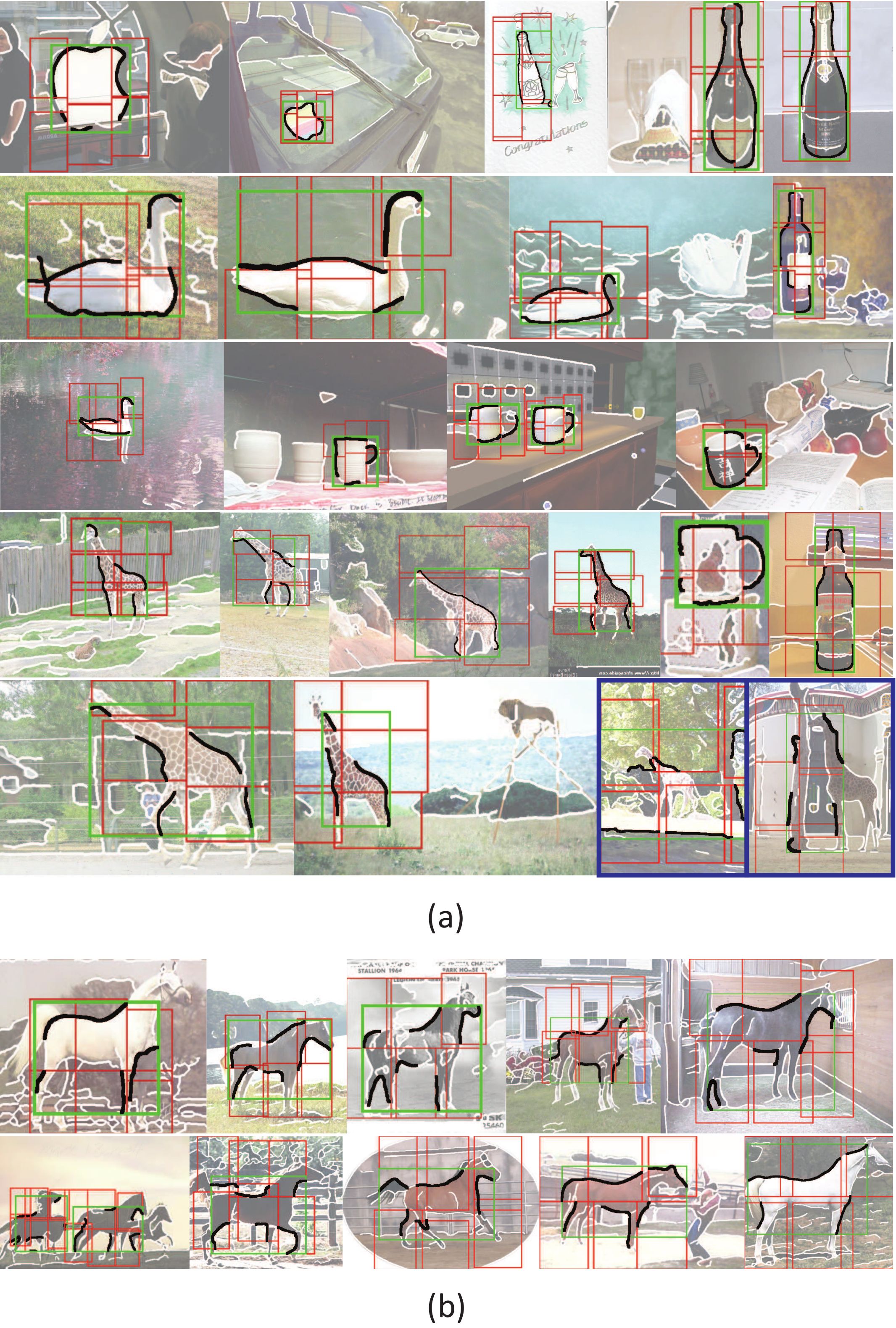,width=3.3in}
\vspace{-5mm}\caption{ A few representative shape detection results generated by our method. Two false positives in (a) are labeled by the bold blue frames.}\label{fig:detect_results}
\end{figure}

{\bf Experiment I.} We use all five classes of shapes from the ETHZ database, (i.e., Apples, Bottles, Giraffes, Mugs and Swans). There are $32 \sim 87$ images in each class, and each image includes at least one shape instance. In the experiments, half of images for each category are randomly selected as positive examples, and a comparative number of negative examples ($70 \sim 90$) extracted from the remaining categories or backgrounds.  The trained shape models for each category are tested on the remaining images. A few typical experimental results are shown in Fig.~\ref{fig:detect_results} (a). For quantitative evaluation, we adopt the Precision-Recall (PR) curves and the average precision (AP) as benchmark metrics, and compare with the state-of-the-art methods~\cite{MalikCVPR2009,ShiShapeCVPR2010,ShapeTree,ShapeGroup,LateckiCVPR2011}. The quantitative results are reported in Fig.~\ref{fig:ETHZ_PR} and in Table ~\ref{tab:ETHZ_AP}. Our method outperforms on $4$ categories (i.e. Apples/Mugs/Giraffes/Swans) which have relatively large intraclass variance or complex backgrounds.

\begin{figure}[!htb]
\centering
\epsfig{figure=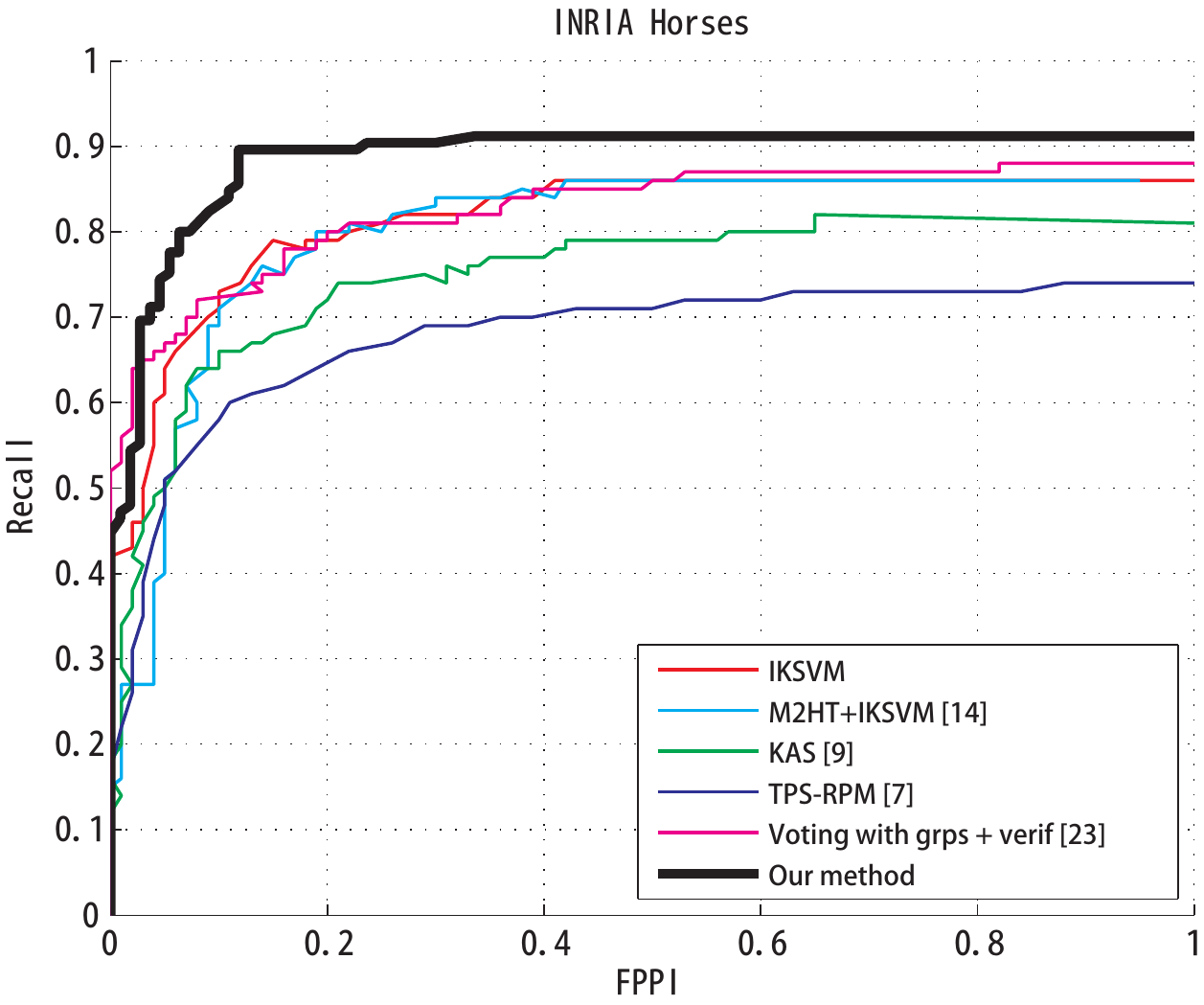,width=2.2in}
\vspace{-2mm}\caption{ Experimental results with the recall-FPPI measurement on the INRIA-Horse database.}\label{fig:Horse_fppi}
\end{figure}

{\bf Experiment II.} The INRIA-Horse dataset consists of $170$ images with one or more horses and $170$ images without horses, which is more challenging than the ETHZ database. Horses appear in images at several scales, and against occlusions and cluttered backgrounds. We randomly select $50$ positive examples and $80$ negative examples for training and test on the remaining images. Fig.~\ref{fig:Horse_fppi} reports the recall against the number of false-positives averaged over all $210$ test images (FPPI). Compared with the recently proposed methods, our system substantially performs better: we achieve a detection rate of $91.2\%$ at $1.0$ FPPI; the reported results of the competing algorithms are: $87.3\%$ in ~\cite{VotingECCV2010}, $85.27\%$ in~\cite{MalikCVPR2009}, $80.77\%$ in~\cite{PAS}, and $73.75\%$ in~\cite{FerrariCVPR07}. From the results of shape detection, some of them are exhibited in Fig.~\ref{fig:detect_results} (b), the improvements are basically made by the accurate location in the context of (i) inconsistent shape contours (caused by pose variants or occlusions) and (ii) noisy edge maps.

\vspace{-1mm}
\section{Summary}
\vspace{-1mm}

This paper studies a novel contour-fragment-based shape model with the And-Or tree representation. This model extends the traditional hierarchical tree structures by introducing the or-nodes that explicitly specify production rules to capture shape variations. Our approach achieves the state-of-art of shape detection on the ETHZ and INRIA-Horse databases. Moreover, the algorithm of And-Or tree learning is very general and can be applied to other vision tasks.


%
%
%
%
%
%
%
%

\bibliographystyle{ieee}

\begin{thebibliography}{\small}

\bibitem{CuttingPlane}
Y. Altun, I. Tsochantaridis, and T. Hofmann, Hidden markov support vector machines, In {\em ICML}, 2003.
\vspace{-1mm}

\bibitem{BaiShapeBand}
X. Bai, Q. Li, L. J. Latecki, W. Liu, and Z. Tu, Shape band: A deformable object detection approach, In {\em CVPR}, 2009.
\vspace{-1mm}

\bibitem{ShapeContext}
S. Belongie, J. Malik, and J. Puzicha, Shape Matching and Object Recognition using Shape Contexts, {\em IEEE TPAMI}, 24(1): 705-522, 2002.
\vspace{-1mm}


\bibitem{DrawingShape2006}
G. Elidan, G. Heitz, and D. Koller, Learning object shape: From drawings to images, In {\em CVPR}, 2006.
\vspace{-1mm}

\bibitem{LatentSVM}
P. Felzenszwalb, R. Girshick, D. McAllester, and D. Ramanan, Object Detection with Discriminatively Trained Part-based Models, {\em IEEE TPAMI}, 2010.
\vspace{-1mm}

\bibitem{ShapeTree}
P. Felzenszwalb, and J. D. Schwartz, Hierarchical Matching of Deformable Shapes, In {\em CVPR}, 2007.
\vspace{-1mm}

\bibitem{FerrariCVPR07}
V. Ferrari, F. Jurie, C. Schmid, Accurate Object Detection with Deformable Shape Models Learnt from Images, In {\em CVPR}, 2007.
\vspace{-1mm}


\bibitem{PAS}
V. Ferrari, L. Fevrier, F. Jerie, and C. Schmid, Groups of Adjacent Contour Segments for Object Detection, {\em IEEE TPAMI}, 30(1): 36-51, 2008.
\vspace{-1mm}


\bibitem{INRIAHorse}
F. Jurie and C. Schmid, Scale-invariant Shape Features for Recognition of Object Categories, In {\em CVPR}, 2004.
\vspace{-1mm}

\bibitem{LinGraphMatch}
L. Lin, X. Liu, and S.C. Zhu, Layered Graph Matching with Composite Cluster Sampling, {\em IEEE TPAMI}, 32(8): 1426-1442, 2010.
\vspace{-1mm}

\bibitem{LinICCV07}
L. Lin, S. Peng, J. Porway, S.C. Zhu, and Y. Wang, An Empirical Study of
Object Category Recognition: Sequential Testing with Generalized Samples, In {\em ICCV}, 2007.
\vspace{-1mm}


\bibitem{LinPR}
L. Lin, T. Wu, J. Porway and Z. Xu, A Stochastic Graph Grammar for Compositional Object Representation and Recognition, {\em Pattern Recognition}, 42(7): 1297-1307, 2009.
\vspace{-1mm}

\bibitem{ShapeGroup}
C. Lu, L. J. Latecki, N. Adluru, X. Yang, and H. Ling, Shape Guided Contour Grouping with Particle Filters, In {\em ICCV}, 2009.
\vspace{-1mm}

\bibitem{LinECCV2010}
P. Luo, L. Lin, and H. Chao, Learning Shape Detector by Quantizing Curve Segments with Multiple Distance Metrics, In {\em ECCV}, 2010.
\vspace{-1mm}

\bibitem{LateckiCVPR2011}
T. Ma and L. J. Latecki, From Partial Shape Matching through Local Deformation to Robust Global Shape Similarity for Object Detection, In {\em CVPR}, 2011.
\vspace{-1mm}

\bibitem{PbDetector}
D. Martin, C. Fowlkes, and J. Malik, Learning to detect natural image boundaries using local brightness, color, and texture cues, {\em IEEET PAMI}, 26(5): 530-549, 2004.
\vspace{-1mm}

\bibitem{MalikCVPR2009}
S. Maji and J. Malik, Object Detection using a Max-Margin Hough Transform, In {\em CVPR}, 2009.
\vspace{-1mm}



\bibitem{SMO}
J. C. Platt, Using analytic qp and sparseness to speed training of support vector machines, In {\em NIPS}, 1998.
\vspace{-1mm}



\bibitem{PartialMatchingECCV2010}
H. Riemenschneider, M. Donoser, and H. Bischof, Using Partial Edge Contour Matches for Efficient Object Category Localization, In {\em ECCV}, 2009.
\vspace{-1mm}

\bibitem{ShottonPAMI08}
J. Shotton, A. Blake, and R. Cipolla, Multi-Scale Categorical Object Recognition Using Contour Fragments, {\em IEEE TPAMI}, 30(7): 1270-1281, 2008.
\vspace{-1mm}

\bibitem{ShiShapeCVPR2010}
P. Srinivasan, Q. Zhu, and J. Shi, Many-to-one Contour Matching for Describing and Discriminating Object Shape, In {\em CVPR}, 2010.
\vspace{-1mm}




%

\bibitem{ConFlexibility}
C. Xu, J. Liu, and X. Tang, 2D Shape Matching by Contour Flexibility, {\em IEEE TPAMI}, 2009.
\vspace{-1mm}

\bibitem{VotingECCV2010}
P. Yarlagadda, A. Monroy and B. Ommer, Voting by Grouping Dependent Parts, In {\em ECCV}, 2010.
\vspace{-1mm}

\bibitem{SVMICML2009}
C.-N. J. Yu and T. Joachims. Learning structural svms with latent variables, In {\em ICML}, 2009.
\vspace{-1mm}


\bibitem{LeoCCCP}
L. Zhu, Y. Chen, A. Yuille, and W. Freeman, Latent Hierarchical Structural Learning for Object Detection, In {\em CVPR}, 2010.
\vspace{-1mm}

\bibitem{ShiShapeECCV2008}
Q. Zhu, L. Wang, Y. Wu, and J. Shi, Contour Context Selection for Object Detection: A Set-to-Set Contour Matching Approach, In {\em ECCV}, 2008.
\vspace{-1mm}

\end{thebibliography}

\end{document}